\documentclass[final]{elsarticle}

\usepackage{hyperref}
\usepackage[shortlabels]{enumitem}
\usepackage{booktabs}
\usepackage{bm}
\usepackage{amsmath,amssymb}
\usepackage[ruled,vlined]{algorithm2e}
\usepackage{algorithmic}
\usepackage{graphicx}
\usepackage{caption}
\usepackage{subcaption}
\usepackage{xcolor}
\usepackage{makecell}
\usepackage{multirow}
\usepackage[margin=3.3cm]{geometry}

\makeatletter
\def\ps@pprintTitle{%
	\let\@oddhead\@empty
	\let\@evenhead\@empty
	\def\@oddfoot{}%
	\let\@evenfoot\@oddfoot}
\makeatother
\begin{document}

\begin{frontmatter}

\title{The Role of Cross-Silo Federated Learning in Facilitating Data Sharing in the Agri-Food Sector}


\author[Glasgow,Aberdeen]{Aiden Durrant}
\author[Aberdeen]{Milan Markovic}
\author[Upton]{David Matthews}
\author[Lincoln]{David May}
\author[Glasgow]{Jessica Enright}
\author[Aberdeen]{Georgios Leontidis\corref{corr}}
\ead{georgios.leontidis@abdn.ac.uk}

\address[Glasgow]{School of Computing Science, University of Glasgow, G12 8RZ, Glasgow,
United Kingdom}
\address[Aberdeen]{Department of Computing Science, University of Aberdeen, AB24 3UE, Aberdeen, \\ United Kingdom}
\address[Lincoln]{Lincoln Institute for Agri-food Technology, University of Lincoln,
LN2 2LG, Lincoln, United Kingdom}
\address[Upton]{Upton Beach Consulting Limited}

\cortext[corr]{Corresponding author}
\begin{abstract}
Data sharing remains a major hindering factor when it comes to adopting emerging AI technologies in general, but particularly in the agri-food sector. Protectiveness of data is natural in this setting: data is a  precious commodity for data owners, which if used properly can provide them with useful insights on operations and processes leading to a competitive advantage.  Unfortunately, novel AI technologies often require large amounts of training data in order to perform well, something that in many scenarios is unrealistic. However, recent machine learning advances, e.g. federated learning and privacy-preserving technologies, can offer a solution to this issue via providing the infrastructure and underpinning technologies needed to use data from various sources to train models without ever sharing the raw data themselves. In this paper, we propose a technical solution based on federated learning that uses decentralized data, (i.e. data that are not exchanged or shared but remain with the owners) to develop a cross-silo machine learning model that facilitates data sharing across supply chains. We focus our data sharing proposition on improving production optimization through soybean yield prediction, and provide potential use-cases that such methods can assist in other problem settings. Our results demonstrate that our approach not only performs better than each of the models trained on an individual data source, but also that data sharing in the agri-food sector can be enabled via alternatives to data exchange, whilst also helping to adopt emerging machine learning technologies to boost productivity.
\end{abstract}


\end{frontmatter}


\section{Introduction}
The agri-food supply chain is a complex and highly valuable sector in the world economy, yet the hostility that arises from competitive advantage has snuffed the possibility of collaboration and openness in data sharing that has the potential to benefit all parties~\cite{ODIsharingsurvey,durrant28might}. Data sharing can help address historical failings related to transparency and traceability of adulterated or unsafe food vertically through the supply chain \cite{sarpong2014traceability}. Substantial work has been done to address the traceability of food and drink with added pressure from consumer demands \cite{pearson2019distributed}. However, in this work we address the data sharing horizontally across the supply chain, as to assist in production optimization or regulatory reporting with the aim to contribute towards the recent international commitments and ambitious goals for sustainability imposed throughout the agri-food supply chain \cite{eusustain}. Specifically, we focus on the implementation of data driven technologies such as machine learning, to gain a statistical insight into a holistic view of data from multiple sources. Moreover, we do so via our proposition for technological methods that facilitate trustworthy data sharing providing a holistic view of a system for optimization or regulatory purposes.

Many of the thorniest challenges of data sharing within agri-food arise from social concerns, perhaps foremost concerns around commercial sensitivity and the resulting reluctance to share, fearing competitive and reputational risks \cite{van2020trust}.  A reluctance to share data may be preventing sector gains from data analytic methodologies (e.g. machine learning) that would allow improved production optimization and environmental data analysis. This fundamental hurdle is the motivation of our work, asking the question: Can we propose technological solutions to facilitate confidence in information sharing within the agri-food sector? Specifically, can we maintain independent ownership of actors' data whilst employing data analysis methodologies to improve production optimization between all actors? 

Production optimization across competitors' activities is an obvious example of the potential gains to be made through analyzing shared agri-food data. Training predictive models on a greater quantity and variety of data from different data owners is likely to produce a more generalizable and better performing model than separate models produced individually by each data owner on their own data. Our goal is to consider methods that can improve on separate, individual models while still not disclosing individual sensitive data. We demonstrate our solution to this goal through
the task of soybean yield prediction, where individual models refers to a model trained on a specific subset of the data belonging to that particular organization only, in our case between US states. The ultimate aim is to move toward the analytic benefits of models trained on pooled data, while avoiding the data pooling itself.


We focus our investigation around the use of data driven technologies across the agri-food supply chain (\textit{i.e.} \emph{horizontally}, rather than \emph{vertically} through the chain given the already vast exploration into this problem setting where the vertical integration has typically been addressed by blockchain technologies \cite{wingreen2019blockchain}). We aim to provide technological solutions to enable trustworthy information sharing among participants across the supply chain that have the ability to provide optimization improvements to support environmental and regulatory concerns. In particular, we trial federated learning (FL) approach and ensemble of models via model sharing to encapsulate many of the established methods in machine learning for agri-food, while not requiring direct data-pooling between competitors. 

FL is an approach to machine learning~\cite{xu2021federated}, which involves training a centralized model collaboratively through many clients whilst keeping client data decentralized \cite{mcmahan2017communication}. Thus, the decentralization of data promotes privacy preservation between individual client data whilst producing trained models that leverage all data of all participating clients. We argue that this setting, specifically the `cross-silo' (few clients that each represent a larger repository of data) provides a potential opportunity to address the challenges faced with distrust in data sharing, producing cooperatively shared models whilst maintaining data independence. Federated learning alone does not prevent all malicious attacks, specifically inference attacks \cite{shokri2017membership} are a significant concern, however to build trust we explore and subsequently demonstrate how the proven theoretical privacy can be achieved in the federated setting through the adoption of differential privacy methods at a participant level. We hope the proposition and initial empirical demonstration of such technologies in the agri-food setting supporting confidence in privacy-preserving methods for information sharing, sparking change in the perceptions of data sharing to build a more sustainable future for agri-food. 


To showcase the potential of FL, ensembles of shared models and differential privacy to facilitate trust for data sharing in agri-food setting, we employ well established open-source datasets for crop yield prediction from both imaging (remote sensing)~\cite{you2017deep} and tabular (weather and soil data)~\cite{khaki2020cnn} data domains. We demonstrate how data independence and privacy preserving methods perform compared to their traditional machine learning counterparts, with the aim to empirically illustrate that under the restricted conditions of independence and differential privacy these training regimes can produce competitive models. In summary this work describes the following contributions:
\begin{enumerate}
    \item We demonstrate the applicability of federated and model sharing machine learning methodologies to enable training of distributed datasets in the settings relevant to the agri-food domain.
    \item We show that the necessary privacy and security concerns prevalent in the agri-food sector can be appropriately overcome via privacy preserving methods, in our case differential privacy.
    \item We argue for the potential adoption of the proposed technological methods to facilitate data sharing, and give key example use cases where such facilitation would benefit all participants.
\end{enumerate}


\subsection{Use cases}\label{usecases}
As our aim is to propose technological solutions to facilitate and subsequently begin to build confidence in data sharing, as well as to potentially encourage those in agri-food to participate, we provide example use cases where we see data sharing in agri-food via distributed training to be most applicable. For this work we primarily focus on production optimization for collaborative federations for our empirical demonstration given the accessibility to open-source datasets. However, the advocated training procedures are directly applicable to the other use cases. We give two key use case problems observed in the agri-food sector that we believe data sharing and collaborative training can assist, this list is not exhaustive.

\begin{itemize}
    \item \textit{Soft fruit production optimization for collaborative federations.} \\
        The addition of more data from a variety of sources and multiple farms vastly improves performance of data driven technologies. In the agri-food sector, soft fruit growers for example, can in some cases be limited by their data collection resources, yet they may wish to employ data analytic to improve not only profits but also their sustainability (net-zero targets). Another angle to this relates to contractual agreements between growers and large retail supermarkets for instance; over- or under- estimating the amount of produce can lead to fruit waste or fruit shortages respectively, that can have both financial and environmental repercussions for growers and the whole sector. The concept of federations like that explored in \cite{durrant28might} facilitated through our proposed federated learning procedure can enable multiple growers to share data in a trustworthy manner to improve their own production systems.
    \item \textit{Analysis of client production from a distribution source - fresh food distribution centre.}\\
        In many cases statistical analysis of a product distributed to clients is beneficial to the supplier. This in turn benefits all participants as data analytical approaches can lead to product improvement tailored to the clients. Like with production optimization, a federated learning setting can allow the central sever (i.e. distribution source) to coordinate training to produce a global statistical model of all their clients so as to gain insights from a holistic view of the data. In the fresh food industry, such an approach can optimize food availability, reduce transport costs, reduce food waste, and contribute to the financial prosperity of all actors involved.
   \end{itemize}

\section{Related Work}

Data sharing has not been as widespread in the agri-food sector as in other domains, such as medical and genomic research, where the realization of how data sharing can improve and accelerate scientific research \cite{milham2018assessment} has outweighed commercial gain of maintaining local data \cite{byrd2020responsible}. While work has taken place in tackling issues of transparency and traceability through technologies such as blockchain \cite{feng2020applying}, true sharing (direct sharing of raw, unprocessed data between parties) of data between actors is typically uncommon within organizational settings. Despite continued support from EU governments for data sharing in agriculture \cite{van2020trust}, recent studies examining the regulatory framework of agricultural data sharing have highlighted the significant extent of the social and technological challenges  \cite{durrant28might}.

The use of machine learning within agri-food has seen successful application in yield prediction \cite{alhnaity2019using}, crop disease detection \cite{shruthi2019review}, and production safety \cite{thota2020multi,onoufriou2019nemesyst}. We suggest that many of these already impactful advancements can be replicated (or at least approximated) without direct data pooling, thus making them much more widely usable. To demonstrate our intuition under the setting of multiple, independent data silos, we give an example of model sharing and decentralized training via federated learning, which has not yet seen widespread application in agri-food.

FL trains machine learning algorithms in a decentralized manner, maintaining principles of focused collection, data minimization, and mitigation of systemic privacy risks associated from traditional centralized methods \cite{kairouz2019advances}. Introduced in 2017 \cite{mcmahan2017communication}, FL has since seen adoption in mobile device infrastructure from tech giants\cite{yang2018applied} and IoT networks. Recently, federated approaches have moved into the mainstream with primary research on extremely large scale `cross-device' settings with millions of edge devices (clients), and continued privacy-preserving implementations. In contrast, our work focuses on the `cross-silo' setting where there are few clients each of which represents a larger data store - this setting is more representative of individual companies or organizations \cite{kairouz2019advances}.

At the core of FL, privacy preserving mechanisms enable the facilitation of confident and trustworthy data mining between independent and decentralized data stores. Under the cross-silo setting there is typically less interest in protecting data from the public domain given the models are generally only released to those who participate in training, and as such more emphasis is placed on inter-client privacy. Although many of the privacy preservation schemes explored defend from such public attacks, while secure communication pipelines like secure aggregation~\cite{bonawitz2017practical} can help reduce such risks. One extensively explored approach is differential privacy \cite{dwork2014algorithmic}, a methodology introducing uncertainty into the released models as to sufficiently mask the contributions of individual data and as such, limit the information disclosure about individual clients. Lately, more emphasis has been placed on more theoretically secure methods of privacy preserving, such as fully homomorphic encryption (FHE), performing computational operations on encrypted data without first decrypting it \cite{gentry2011implementing, onoufriou2021edlaas}.

\section{Problem Setting}\label{problemsetting}

Traditionally, training large statistical models to provide a holistic analysis of data requires the collection of many data points typically originating from various independent collections. Informally, this can be considered as bringing the data to the model for training, pooling multiple data silos (independent datasets/databases belonging to an single client or participant \textit{e.g.} organization, county, nation) into a centralized data store. However, as mentioned this unification and direct data sharing is deemed impractical in the agri-food sector due to privacy concerns, distrust and subsequent risk to commercial sensitivity \cite{durrant28might}. One particular example implication being addressed by distributed training is the direct analysis of yield information and the subsequent inference of sensitive financial information. Removing the ability of participants to access other participant data mitigates such trivial attacks. Following such an example we ask, \textit{can we train machine learning models that leverage data from many individual data silos without explicitly sharing or centralizing data?} This in-turn focuses our investigation to first undertake the task of distributed training on multiple independent data silos, and secondly ensure data privacy is maintained through appropriate mechanisms to elicit trust.

To tackle distributed training we explore model sharing, transferring independently trained models rather than data itself (Section \ref{modelsharing}), and federated learning for training an aggregated model on multiple independent data silos simultaneously (Section \ref{federated}). This scenario represents the case whereby clients want to protect their own data, hence being reluctant to share them with other growers. The current availability of real-world data belonging to many different organizations or groups of the same type is limited within the agri-food setting, and thus we simulate the case of multiple, independent, distributed data-stores by partitioning two well established, open-source datasets comprised of five subsets~\cite{DAAC2015,USDA-NASS2019,Daymet2016,gSSURGO2019} into sub-sets each representing a separate data silo.

To adequately compare the performance of our proposed approaches, and to empirically support our agenda of augmenting traditional machine learning with distributed learning algorithms in agri-food, we employ the two benchmark datasets aforementioned to measure performance. Both datasets address the task of yield prediction of soybean production in the US corn belt region, one utilizing remote sensing data from satellite imagery and the other with more traditional tabular data corresponding to soil conditions, weather, etc. County-level predictions are made given this is the granularity of the dataset in question.

\subsection{Remote Sensing Yield Prediction}

The first of our datasets for empirical analysis focuses on the well established problem of average yield prediction of soybean from sequences of remote satellite images taken before harvest \cite{you2017deep}. More concretely, we focus on prediction of the average yield per unit area within specific geographic boundaries, \textit{i.e.} counties of 11 US states in the corn belt. The sequence of remote images in question are multi-spectral images taken by the Terra satellite, with each image $(\bm{I}^{(1)},\dots,\bm{I}^{(T)})$ corresponding to a county region. The sequence is temporal with readings taken at equally-spaced intervals 30 times throughout the year ($T=30$), $\bm{I}^{(t)}$ represents the image at time $t$ within a year. 

Our goal remains the same as described in \cite{you2017deep}, to map the raw multi-spectral image sequences that capture features related to plant growth to predict average observed yield, the difference here relates to the training setting and dataset structure to allow for fair comparisons and evaluation of our training procedures. In our case we aim to learn a model trained on multiple, independent data silos, synthetically generated by splitting the dataset $D$ per US state as to produce 11 data silos. The resulting dataset per state silo $k$ is given by
\begin{multline}
    D_k = \bigg\{ \left( (\bm{I}^{(1)}_k, \dots, \bm{I}^{(T)}_k, g_k^{\mathtt{loc}},g_k^{\mathtt{year}})^{[1]}, y_k^{[1]} \right), \dots, \\ 
    \left( (\bm{I}^{(1)}_k, \dots, \bm{I}^{(T)}_k, g_k^{\mathtt{loc}},g_k^{\mathtt{year}})^{[N]}, y_k^{[N]} \right) \bigg\}
\label{eq:contrastive}
\end{multline}

where $g_{\mathtt{loc}}$ and $g_k^{\mathtt{year}}$ are the geographic location and harvest year respectively, $y \in \mathbb{R}^+$ is the ground truth crop yields, and $N$ is the number of data samples in $D_k$. Statistics are given in Table \ref{tab:sample_stats_img} outlining the split samples following this procedure and the average yield for the test set given the prediction year of 2016.

Furthermore, the images $\bm{I}$ are transformed into histograms of discrete pixel counts to reduce the dimensional of the satellite images that make training machine learning systems with a relatively small dataset challenging. A separate histogram is constructed from each imaging band (in our case $d=9$ bands) and these are concatenated to form $\bm{H} = (\bm{h}_1,\cdots,\bm{h}_d)$, where for each time step $\bm{H}^{(t)}\in \mathcal{R}^{b \times d}$, this is the input into our networks.

\begin{table}[h]
    \scriptsize
    \centering
    \begin{tabular}{c|c|c|c|c||c}
        \multirow{ 2}{*}{State / Silo} & Number of & Train & Valid & Test & Average Test Yield\\
         & Counties & Samples & Samples & Samples & (Bushels/Acre)\\
        \midrule
        Arkansas & 29 & 404 & 27 & 44 & 39.20\\
        Illinois & 97 & 1151 & 93 & 127 & 49.01\\
        Indiana & 86 & 1012 & 82 & 112 & 48.55\\
        Iowa & 99 & 1157 & 99 & 128 & 49.93\\
        Kansas & 68 & 912 & 59 & 101 & 35.97\\
        Minnesota & 73 & 859 & 71 & 95 & 41.86\\
        Missouri & 75 & 981 & 66 & 108 & 38.41\\
        Nebraska & 70 & 857 & 66 & 95 & 51.87\\
        North Dakota & 41 & 383 & 35 & 42 & 30.79\\
        Ohio & 78 & 901 & 34 & 100 & 46.59\\
        South Dakota & 45 & 537 & 41 & 58 & 37.49\\
        \midrule
        Combined / Pooled & 9154 & 713 & 1010 & 42.71
    \end{tabular}
    \caption{Remote Sensing data sample breakdown for the silo-ed setting where the silos coincide to each state. The validation set corresponds to the year before the prediction year, and the test set corresponds to the prediction year, 2016.}
    \label{tab:sample_stats_img}
\end{table}

\subsection{Tabular Yield Prediction}

To further demonstrate the performance of the proposed methods under different modalities other than images, we also employ tabular data for the same task of soybean yield prediction, although for a smaller geographical area comprising of 9 US states and their counties. As with the remote sensing data, this dataset also aims to predict the observed average yield of soybean. The features used to map to the average yield are as follows:

\begin{itemize}
    \item \textit{Crop Management:} In addition to the yield performance (our prediction target) we also use the weekly cumulative percentage of planted fields within each state, starting from April each year, as indication of planting time. The crop management data were obtained from the public domain from the National Agricultural Statistics Service of the United States \cite{USDA-NASS2019}.
    \item \textit{Weather Components:} Weather data have been acquired from the Daymet service \cite{Daymet2016}, providing daily records of weather variables including: precipitation, solar radiation, snow water equivalent, maximum temperature, minimum temperature, and vapor pressure. The resolution of each data variable is 1 $\text{km}^2$.
    \item \textit{Soil Components:} 11 soil variables are measures for 6 depths 0-5, 5-10, 10-15, 15-30, 30-45, 45-60, 60-80, 80-100, and 100-120 cm at a spatial resolution of 250 $\text{m}^2$. The 11 soil components are: soil bulk density, cation exchange capacity at pH7, percentage of coarse fragments, clay percentage, total nitrogen, organic carbon density, organic carbon stock, water pH, sand percentage, silt percentage, and soil organic carbon. This data is provided by Gridded Soil Survey Geographic Database for the United States \cite{gSSURGO2019}, measured at a spatial resolution of 250 $\text{km}^2$.
\end{itemize}

The data is organized by year, per county for the years 1980 to 2018, with each county's average yield being given alongside the planting date, soil components, and the weather variables measured weekly for that year. The data had been collected, cleaned and provided kindly by the authors of \cite{khaki2020cnn}, please refer to their work for further processing and cleaning details. Lastly, to appropriately simulate the setting of multiple, independent data silos, we follow the same procedure as described for the remote sensing data, dividing the complete dataset into 9 subsets, each representing an individual data silo of a US state and their corresponding counties. As with the remote sensing data, statistics are given in Table \ref{tab:sample_stats_tab} outlining the split samples per state silo and the average yield for the test set given the prediction year of 2018.

\begin{table}[h]
    \scriptsize
    \centering
    \begin{tabular}{c|c|c|c|c||c}
        \multirow{ 2}{*}{State / Silo} & Number of & Train & Valid & Test & Average Test Yield\\
         & Counties & Samples & Samples & Samples & (Bushels/Acre)\\
        \midrule
        Illinois & 100 & 3480 & 91 & 75 & 56.76 \\
        Indiana & 92 & 3151 & 81 & 74 & 53.74\\
        Iowa & 98 & 3622 & 94 & 91 & 56.41\\
        Kansas & 105 & 3245 & 51 & 39 & 40.74\\
        Minnesota & 84 & 2605 & 66 & 50 & 48.17\\
        Missouri & 113 & 2973 & 61 & 39 & 48.67\\
        Nebraska & 89 & 2787 & 33 & 51 & 57.50\\
        North Dakota & 49 & 943 & 28 & 20 & 34.49\\
        South Dakota & 62 & 1524 & 35 & 33 & 44.71\\
        \midrule
        Combined / Pooled & 792 & 24330 & 540 & 472 & 55.33
    \end{tabular}
    \caption{Tabular data sample breakdown for the silo-ed setting where the silos coincide to each state. The validation set corresponds to the year before the prediction year, and the test set corresponds to the prediction year, 2018.}
    \label{tab:sample_stats_tab}
\end{table}

\section{Model Sharing} \label{modelsharing}
In the pursuit of data independence, we first ask the question if data sharing is even necessary at all, and instead consider the possibility of sharing only trained models among participants. This concept that we term `model sharing' enables each participant $k$ to train a machine learning model on their own data silo $D_k$ independently and distribute these trained models. The fundamental principles of this concept have historically found great success in deep learning, with transfer learning~\cite{cao2010adaptive} and domain adaptation~\cite{wang2018deep} enabling the transfer of learned knowledge in one setting to be exploited to improve generalization in another setting. Yet in the setting of many models transfer learning becomes impractical due to fine-tuning and training difficulties.

From this notion and focusing on our problem setting of yield prediction, we first explore how to leverage each model trained on a particular participants data, to enable a prediction under a holistic view of the data. The problem of how to leverage multiple statistical models is a well studied one~\cite{dong2020survey}, yet is vastly dependent on task, setting, data, and model architecture. Inspired by works such as \cite{kamnitsas2017ensembles} we explore how the use of ensemble predictors of multiple models can be used to leverage the features across all independent models as a collective. 

\begin{figure}[h!]
    \centering
    \includegraphics[scale=0.3]{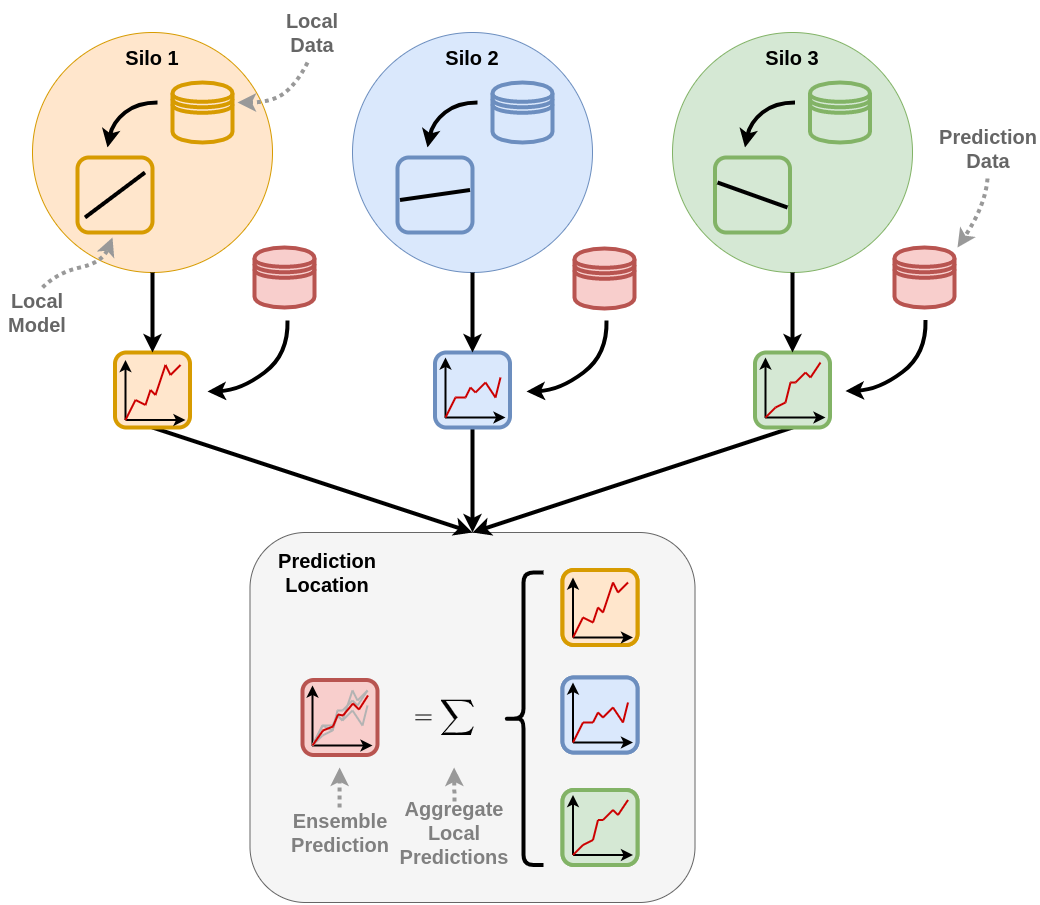}
    \caption{Visual depiction of the ensemble model sharing methodology. Each client trains a model on their own local data, then this model is evaluated on some prediction data, the resulting predictions are aggregated to produce a global prediction.}
    \label{fig:modelshare}
\end{figure}

Fundamentally, ensembles allow for improved generalization to unseen data by removing the need for complex human designed heuristics to choose which model may be most appropriate for a given prediction on unseen data. Specifically in our case, each model $F_k$ and its parameters $\theta_k$ is trained on a unique subset of data $D_k$ that contains its own feature distributions, and as such there is no guarantee that a single model prediction will perform well. The aggregation of multiple model predictions ensures that this single poor model performance is not the sole prediction rather the predictions from all feature distributions and data domains is employed. Formally, the aggregation procedure for the prediction $y$ on data point $x$ can be written as follows
\begin{equation}
P(y|x) = \frac{1}{|K|}\sum_{\forall k \in K} P(y|x;\theta_k)
\label{eq:ms}
\end{equation}
where $\theta_k$ is the parameters of the machine learning model trained on the data $D_k$ from participant $k$, and $|K|$ represent the number of participants/models in the ensemble. Moreover, this process is visually depicted in Figure \ref{fig:modelshare}.

The simple average aggregations shown in Equation~\ref{eq:ms} gives equal weight to every model prediction, however, in reality some models may be trained on data that is relevant to the unseen test data. In our case we have differences in geographic location that can have significant impact on performance~\cite{you2017deep}, and as such the aggregation should to take this into account. Therefore we introduce a weighted averaging scheme that simply weights the predictions by geographical location (furthest distance, lowest weight). This extends Equation~\ref{eq:ms} and weights each prediction by its distance ranking from the location of the prediction data to the location of the data that model has been trained on. The weights are defined as follows
\begin{equation}
W_k = d(g^{loc}_x, g^{loc}_k)
\label{eq:wms}
\end{equation}
where $d(\cdot, \cdot)$ is the location distance ranking between the test time data $x$ and the training data $k$ (shortest geographical distance $W_k = 1$, longest geographical distance $W_k = |K|$).

Ensemble methods for model sharing provide a mechanism to leverage the knowledge learned from independent machine learning training without the need to share raw data, all whilst enabling prediction with a holistic view of the data supporting our potential use cases in Section \ref{usecases}. This method although simple addresses the concern of data sharing and trust with the idea of data independence and the sharing of less interpretable information. We empirically demonstrate the performance of such a method to showcase its potential in agri-food in Section~\ref{experimental}. Yet security and privacy concerns still exist in the transfer of models due to malicious attempts to extract raw data via methods such as inference attacks \cite{shokri2017membership}. We later explore a potential mitigation to these attacks in Section~\ref{localdiff} with deferentially private learning of the individual models, and also discuss the implication introduced with sharing information such as trust in a central organisation party in Section~\ref{discussion}. 

\section{Federated Learning}\label{federated}

The overarching principles presented by model sharing and the ensemble-based training regime demonstrate the potential for information sharing without disclosing raw data. Although these methods show adequate performance (Section \ref{ms_per}), there exists more appropriate methods to leverage the distributed, independent datasets attributed to our problem setting. We focus now on a natural extension to sharing trained models among participants, instead, training a single model via the simultaneous communication of individual model updates by each participant. Known as federated learning, we aim to solve our machine learning problem defined in Section \ref{problemsetting} collaboratively via multiple participants under the coordination of a central server~\cite{kairouz2019advances} without disclosing or sharing raw data rather sharing model updates (i.e. weights and biases). 

Typically, federated learning is described in a \textit{cross-device} setting where there are potentially millions of participants each with an unique dataset (i.e. IoT~\cite{lu2019blockchain} and mobile phones~\cite{bonawitz2017practical}), however there are also many settings where federated learning can be applicable to a relatively few number of participants~\cite{sheller2020federated}. The latter is known as \textit{cross-silo} federated learning and specifically differs from cross-device by the quantity, size and availability of the participants data. As alluded to in the name, cross-silo federated learning collaboratively trains a shared model on siloed data --- data is partitioned by example and also by features, where in our problem setting the features are independent among participants --- that tends to be almost always available, typical of our problem setting where individual organization's data can be reasonably considered as data silos. We will refer to the cross-silo setting when discussing federated learning throughout this work.

Formally each participant, known as a client, contributes to the training of a single global model coordinated by a central server that minimizes the error over the entire dataset, where this dataset is the union of the data across clients. The process of training a model via federated learning is given as follows, where we describe the \texttt{FederatedAvgeraging} algorithm \cite{mcmahan2017communication}:
\begin{enumerate}
    \item The central server initializes the global model architecture and parameters $w_0$. 
    \item \textit{Start of Round:} $C$ fraction of client silos $K$ are selected, $S$, for the round $t$. The global model is sent via private communication to each of the chosen client silos in $S_t$.
    \item Once received by the client $k$, the global model is trained on the local subset of data belonging to that client silo only. This is a standard gradient descent optimization procedure that results in an updated model referred to as the local model. Each local model per client is an unique model representing that individual silos data.
    \item Following local training, the local model weights are privately communicated back to the central server where they are aggregated and averaged over the individual clients $k$ to produce a new global model $w_{t+1}$.
    \item Steps 2-4 are repeated for the number of communication rounds $T$.
\end{enumerate}

\begin{figure}[h]
\captionsetup[subfigure]{font=small,labelfont=small}
    \centering
    \begin{subfigure}[t]{0.45\textwidth}
      \centering
        \includegraphics[width=0.75\textwidth]{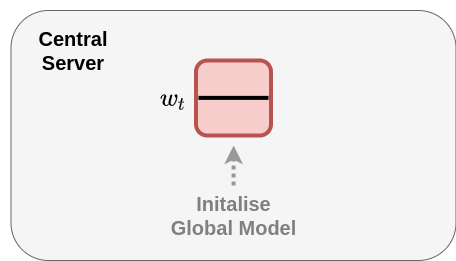}
        \caption{Central node initializes the model paramters.}
    \end{subfigure}\hfill
    \begin{subfigure}[t]{0.45\textwidth}
      \centering
        \includegraphics[width=0.95\textwidth]{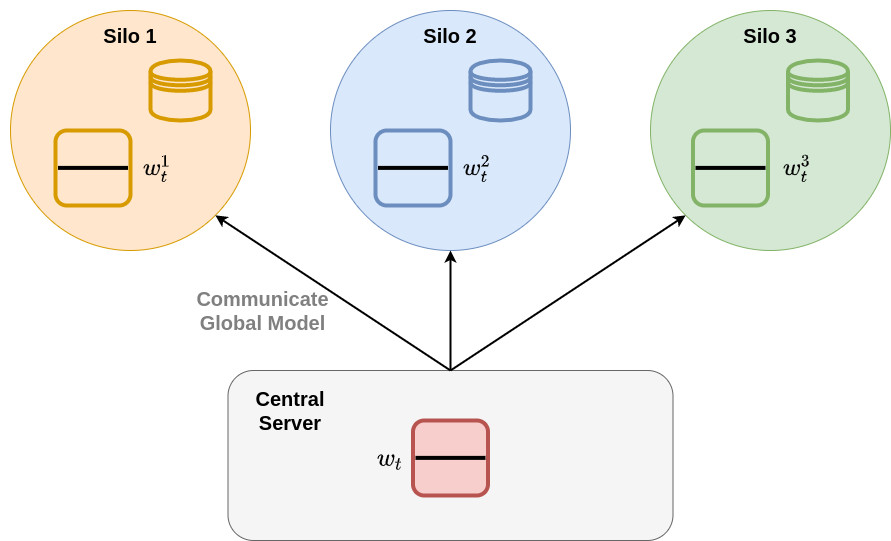}
        \caption{Each client receives the initialized global model from the central server.}
    \end{subfigure}
    
    \begin{subfigure}[t]{0.45\textwidth}
      \centering
        \includegraphics[width=0.99\textwidth]{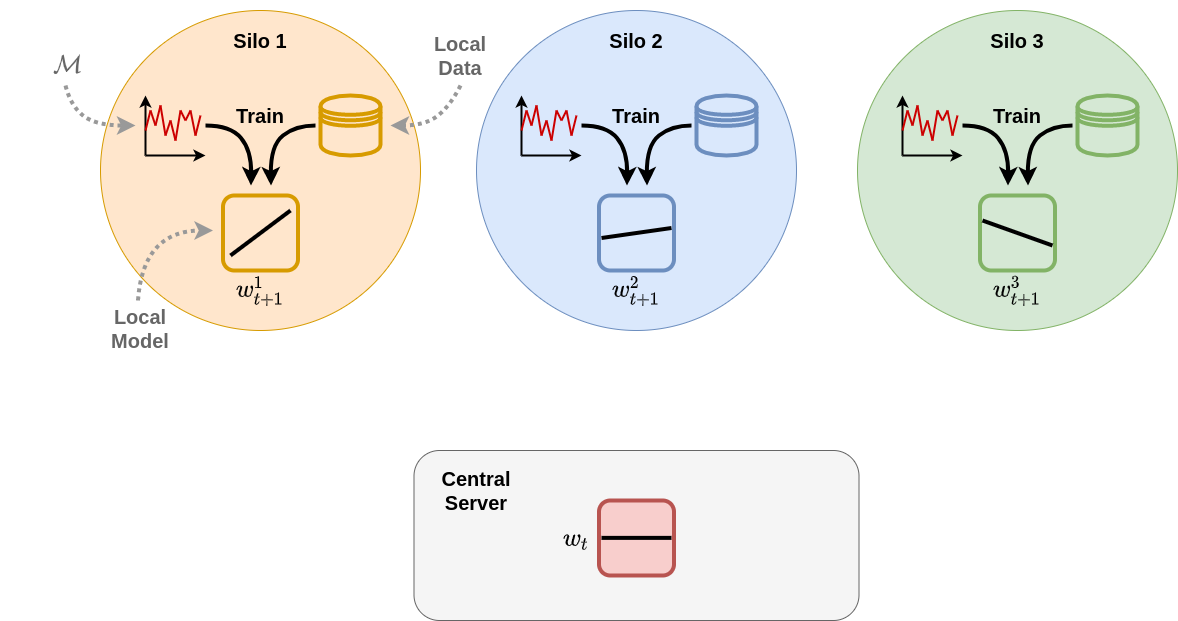}
        \caption{Each client trains its copy of the global model on its own local data to produce an updated local model. In the local deferentially private setting this involves some addition of noise.}
    \end{subfigure}\hfill
    \begin{subfigure}[t]{0.45\textwidth}
      \centering
        \includegraphics[width=0.95\textwidth]{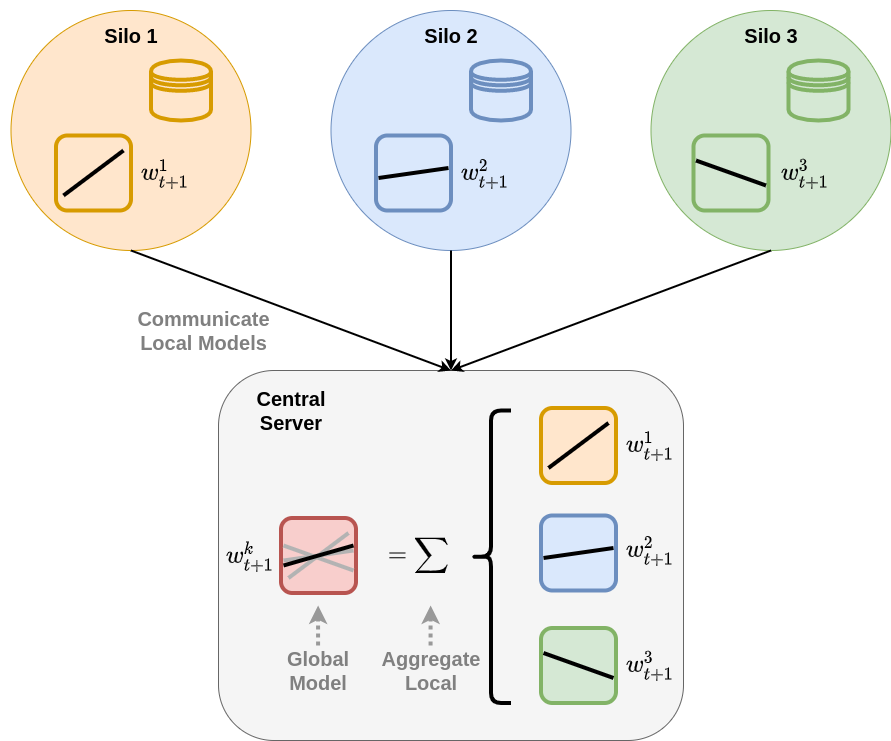}
        \caption{The clients send their local models to the central server where they are aggregated to produce an updated global model. Steps b-d repeat for a number of communication rounds.}
    \end{subfigure}
    \caption{Depiction of centralized cross-silo federated learning, \texttt{FederatedAveraging}. The temporal process moves left to right (a-d).}
\label{fig:fed}
\end{figure}

This process is formally defined in Algorithm \ref{alg:1} and visually depicted in Figure \ref{fig:fed}.

\begin{minipage}{.75\linewidth}
\small
\begin{algorithm}[H]
\SetAlgoLined
\DontPrintSemicolon
    \KwSty{Central Server Executes:} \\
    initialize $w_0$\\
    \ForEach{\normalfont{round} $t=\{1,2,\dots\}$}{
     $m \leftarrow \max (C \cdot K,1)$\\
     $S_t \leftarrow$ \normalfont{(random set of $m$ silos)}  \\  
    \ForEach{\normalfont{silo} $k \in S_t$ \textbf{in parallel}}{
    $w_{t+1}^k \leftarrow$ \normalfont{SiloUpdate($k,w_t$)}\\
    }
    $w_{t+1} \leftarrow \sum^K_{k=1} \frac{n_k}{n} w^k_{t+1}$ \\
    
    }
    \texttt{\\}
    \KwSty{SiloUpdate($k,w$):} \ArgSty{ // On silo $k$} \\
    $\mathcal{B} \leftarrow$ \normalfont{(split $D_k$ into batches of size $B$)} \\
    \ForEach{\normalfont{local epoch} $i=\{1,\dots,E\}$}{ 
    \For{\normalfont{batch} $b \in \mathcal{B}$}{
    $w \leftarrow w - \eta \cdot \mathcal{M}(\triangledown \ell (w;b))$\\
    }
    
    }
   
 \caption{\texttt{DP-FederatedAveraging}. The 
 $K$ silos are indexed by $k$; $C$ is the fraction of silos used per round, $B$ is the local minibatch size, $E$ is the number of local epochs, and $\eta$ is the learning rate. $\mathcal{M}$ is the local differential privacy-compliant algorithm.}
 \label{alg:1}
\end{algorithm}
\end{minipage}

As inferred from the previous definition and description of the federated learning training procedure, explicit privacy advantages can be observed in comparison to traditional machine learning training on centralized and persisted data. Most obvious is the distributed nature of the data held by the clients, maintaining data independence and subsequently addressing issues related to trust and commercial sensitivity of the agri-food sector identified and discussed in \cite{durrant28might}. In addition, it has been well understood how such methods fit into legislative limitations regarding GDPR~\cite{truong2020privacy} further introducing regulatory confidence in such a methodology to maintain data privacy between clients.

Focusing further on our particular problem setting of the agri-food sector, the concept of a central communication server could potentially introduce obstacles that relate to malicious information retrieval. One such approach to address this is to consider the concept of data trusts as presented in \cite{durrant28might, ODI}. This introduces a trusted party to maintain and facilitate the central server communication. Additionally, methods such as differential privacy can further alleviate concerns of malicious attacks on the global and local models obtained via interception of communications or through legitimate means during the sharing procedure between all participants, this is elaborated on in Section \ref{discussion}.

\subsection{Statistical Heterogeneity of Silos} \label{noniid}

The \texttt{FederatedAveraging} algorithm provides the basis for the implementation of many federated learning systems. However, in most real-world cases and including our problem setting, there exist complications regarding data that is not independently and identically distributed (iid). In our particular setting of the agri-food sector, it is likely that \textit{feature shifts} are the most common factor in non-iid data. Informally, a feature shift may result from a difference in local measurement devices or sensitivity of measurements used to obtain the data for each of the local participant data silos. This shift in feature distribution can lead to significant performance degradation as each local model is trained on a distribution that is not aligned with other clients. As a result the global model is averaged across a number of shifted distributions leading to a model that is not appropriately representative or generalisable to the union of the individual local datasets. Formally feature shift is defined by two properties of the probability between features $x$ and labels $y$ on each client: 1) \textit{covariate shift:} the marginal distributions $P_k(x)$ varies across clients, even if $P_k(y|x)$ is the same for all clients; and 2) \textit{concept shift:} the conditional distribution $P_k(x|y)$ varies across clients while $P_k(y)$ remains the same. 

\begin{figure}[h]
    \centering
    \includegraphics[width=0.64\textwidth]{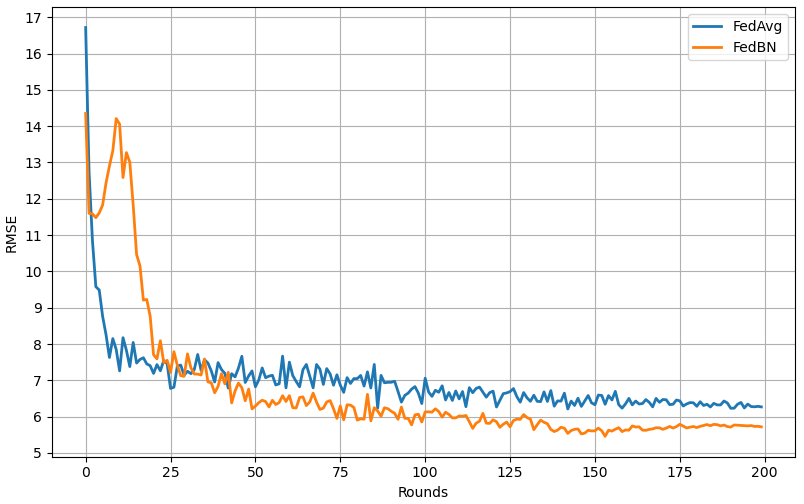}
    \caption{RMSE (Bushels per Acre) over all prediction years (CNN model for remote sensing dataset).performance of the federated learning aggregation methods over communication rounds trained and tested on the remote sensing data for the prediction year 2015, 4 epochs per round.}
    \label{fig:rounds}
\end{figure}

To overcome statistical heterogeneity due to feature shift we consider the \texttt{FedBN} (federated batch normalization) algorithm \cite{li2021fedbn} as to extend \texttt{FederatedAveraging}. \texttt{FedBN} first assumes the model to be trained locally contains batch normalization layers, for our problem setting replicating the architectures in ~\cite{khaki2020cnn} this assumption holds true. Informally, \texttt{FedBN} extends \texttt{FederatedAveraging} by simply excluding the batch normalization parameters from the averaging step, instead maintaining the local parameters for each model. However, the statistical heterogeneity demonstrated by each silo results in batch normalization parameters, that control the standardization of layers, being inappropriate for specific subset distributions. This is theoretically demonstrated in \cite{li2021fedbn} and empirically shown in Table \ref{tab:FedMethods} and Figure \ref{fig:rounds} to improve performance by appropriately standardizing the activations for that data silo's distribution.

Alternative approaches address issues in non-iid data specifically focusing on label distribution skew, specifically FedProx~\cite{li2018federated}, and FedMA~\cite{wang2020federated}. The point raised in mitigating the effect of statistical heterogeneity in data silos is vastly important to the implementation and performance of federated learning, and as such should be considered a key aspect in the adoption of such methods in practice. Moreover, the adoption of methods to tackle statistical heterogeneity is dependent on the data itself, and should be considered as important as feature normalization in machine learning.

\begin{table}[h]
    \centering
    \scriptsize
    \begin{tabular}{c|c|c}
        Method & Imaging Dataset & Tabular Dataset \\
        \midrule
        FedAvg & 5.679 & 3.050 \\
        FedBN & 5.593 & 2.782 \\
    \end{tabular}
    \caption{Comparison of federated learning aggregation methods under both dataset modalities, average RMSE (Bushels per Acre) over all prediction years (CNN model for remote sensing dataset).}
    \label{tab:FedMethods}
\end{table}

\section{Local Differential Privacy} \label{localdiff}

The model sharing and distributed training methodologies described address the primary concern outlined in our problem setting, that alternatives to explicit raw data sharing need to be employed to elicit trust in data holders. Both propositions share trained statistical models rather than the data itself that by themselves can be seen as less interpretable and less vulnerable to malicious use. We later discuss in Section~\ref{discussion} potential solutions and procedures to ensure the models themselves are appropriately maintained and distributed to avoid misuse. However, it is beneficial to mitigate malicious attempts before the communication takes place. 

Differential privacy~\cite{dwork2014algorithmic} operates under the notion of uncertainty within the shared models to mask the contribution of any individual user, where for machine learning the ability of what an adversary can learn about the original training data based on analyzing the parameters is severely limited~\cite{mcmahan2017communication}. Formally, a randomized mechanism $\mathcal{M}: \mathcal{D} \rightarrow \mathcal{R}$ with a domain $\mathcal{D}$ (e.g. training datasets) and range $\mathcal{R}$ (e.g. trained models) is ($\epsilon, \delta$)-differentially private if for any two adjacent datasets $d, d' \in \mathcal{D}$ and for any subset of outputs $\mathcal{S} \subseteq \mathcal{R}$ the following equation holds
\begin{equation}
\text{Pr}[\mathcal{M}(d) \in \mathcal{S}] \leq e^\epsilon \text{Pr}[\mathcal{M}(d') \in \mathcal{S}] + \delta.
\label{eq:dp}
\end{equation} 
When we apply this definition to a mechanism $\mathcal{A}$ that processes a single clients local dataset $\mathcal{D}$, with aforementioned guarantee holding with respect to any possible other local dataset $\mathcal{D'}$, we refer to this setting as \textit{local model of differential privacy}~\cite{ding2017collecting, truex2020ldp}. The local aspect ensures that differential privacy is employed at the client level during the training procedure, thus local models communicated hold a level of privacy. This mechanism differs from more standard approaches, where in the federated learning setting a central server is trusted to apply the randomized mechanism, and therefore requires trust in the communication and in the server itself, a significant social challenge in our setting.

We employ the DP-SGD~\cite{geyer2017differentially} algorithm for differential privacy in a local fashion inspired by~\cite{truex2020ldp}, the process of applying our randomized mechanism $\mathcal{M}$ is visually depicted in Figure \ref{fig:fed} and shown in Algorithm \ref{alg:1}. Local differential privacy is employed over its global variant (i.e., applied after the aggregation on the server side) due to reliance on a trustworthy server, in the agri-food setting the preservation of privacy for each model before it leaves the local client to be aggregated to form the global model is optimal. However, the local setting typically performs worse that global due to an increased quantity of noise added to each sample from each client rather than one server side addition. Subsequently, in practice higher values of $\epsilon$ and $\delta$ are employed limiting the privacy guarantee to ensure the quantity of noise added to the samples is not damaging to the performance. As observed in Equation \ref{eq:dp}, the value of $\epsilon$ is the absolute privacy guarantee that you cannot gain an $e^\epsilon$ amount of probabilistic information about a single entry between $d$ and $d'$, whereas $\delta$ is the value which controls failure of differential privacy guarantee. For each entry there is a $\delta$ probability this failure may happen, so in general this will occur $\delta \cdot n$ times, where n s the number of entries. We therefore aim for both $\epsilon$ and $\delta$ to be small if we wish for upmost privacy. However, the value for $\epsilon$ increases throughout training as more passes over the data are made, as such there is a distinct trade-off between privacy budget $\epsilon$ and performance from the number of epochs of training. We empirically demonstrate this on our problem setting and data, with the results depicted in Figure \ref{fig:epsilon}. We show that as we increase the privacy budget reducing the privacy guarantee we also increase the performance, subsequently we empirically decide the performance-privacy trade-off for our problem setting and data.

We provide an experimental study in Section \ref{experimental} regarding optimal parameters of the differential privacy mechanism including $\epsilon$, $\delta$, and noise values, as well as the performance of the yield prediction task under differential privacy guarantees. Importantly, these values that control privacy such as $\epsilon$ are tuneable and/or flexible to the practitioner conducting the training. Additionally, in the local differential privacy settings individual participants can control their own privacy budgets independently thus controlling their privacy to performance trade-off. The flexibility of these approaches is vastly beneficial in agri-food where tasks and goals vary so widely.

\section{Experimental Results} \label{experimental}

The aforementioned paradigms for training machine learning systems in the setting of independent, distributed data silos promote privacy preservation and can potentially facilitate trust and cooperation in sharing information. Although these methods provide strong theoretical and practical guarantees for privacy \cite{dwork2014algorithmic,truex2020ldp}, the performance of the trained machine learning models must still be adequate to solve the tasks at hand. The obvious benefit of more data, naturally improves these data driven optimization procedures, and the view of holistic data analysis is well defined in this work as a potential use case in the agri-food sector. Yet, we provide an empirical study to further validate our propositions in the agri-food sector and demonstrate its applicability to current problems to establish confidence in performance of distributed data driven computation. 

\subsection{Model and Data Description}

We first define the machine learning models and procedures used to perform our task of average soybean yield prediction per county of US states in the corn belt. Specifically we do so under two modalities of data, remote sensing satellite imagery, and soil, weather, and crop management readings (Tabular). As described in Section \ref{problemsetting}, we aim to perform this machine learning training in a distributed and independent manner where the dataset is comprised of local subsets of the data belonging to individual states. The training procedures previously defined utilize the same core machine learning model architectures throughout all experiments for that modality unless mentioned otherwise. 

\subsubsection{Imaging with remote sensing data}

To most appropriately compare and evaluate the performance of our machine learning training paradigms we first explore the well established task of average soybean yield prediction by remote sensing satellite imagery, where the data itself is described in Section~\ref{problemsetting}. Given our work focuses on the exploration of methods to facilitate information sharing we instead replicate the models described in \cite{you2017deep} to provide an established baseline. A convolutional neural network (CNN) consisting of six convolution $\rightarrow$ batch normalization $\rightarrow$ ReLU $\rightarrow$ dropout blocks and one multi-layer perceptron (MLP) is employed, a recurrent neural network (RNN) is also experimented, specifically one long short-term memory layer consisting of 128 units followed by two MLP layers separated with batch normalization and ReLU. Both of these networks are defined and described in \cite{you2017deep}, where for this work we disregard the Gaussian process procedure given its inability to be performed in the distributed manner defined in our problem setting. We train all networks for 160 epochs (or for federated learning 4 epochs locally with 40 communication rounds) yet this may end prematurely due to the use of early stopping. Furthermore we use the stochastic gradient descent (SGD) optimizer with a learning rate of 0.0001 decaying at 60 and 120 epochs by a factor of 0.1, the remainder of hyperparameters are identical to \cite{you2017deep}. 

The performance is reported by the root mean square error (RMSE) of the county-level bushels per acre yield predictions averaged over 3 runs of 3 seeds, where we evaluate the test set per state silo. The predictions are made for 7 years (2009-2015) where for the given year the model is trained on all data from preceding years. We report a baseline performance which refers to the traditional setting of pooling all the data and training one single model. The model sharing ensemble and federated learning setting are trained on local datasets pertaining to the individual states. All models were tuned using a 15\% hold out validation set. 


\subsubsection{Tabular weather, soil and crop management data}

Following the exploration of the remote sensing data we also demonstrate performance on a more traditional tabular dataset in order to show how differing data domains perform under the outlined training paradigms. Importantly, both datasets are employed to perform the same task, although the datasets vary in features and collection. Nevertheless, we utilize the CNN-RNN defined in \cite{khaki2020cnn} as an established baseline to report the performance in the same manner as the remote sensing data, root mean square error (RMSE) of the county-level bushels per acre yield predictions, evaluated over the test set per state silo. However we make some slight changes to the implementation provided by \cite{khaki2020cnn} such as the addition of batch normalization, removal of the MLP layers and we maintain that the CNN layers only operate on a single time-step before being fed to the RNN. This resulted in our adjusted network outperforming that in \cite{khaki2020cnn} on the same data whilst maintaining the core concept presented in their work. We train for 60 epochs (or for federated learning 4 epochs locally with 15 communication rounds) with early stopping via SGD optimization with a learning rate of 0.001 decaying at 20 and 40 epochs by a factor of 0.1, the remainder of hyperparameters are identical to \cite{khaki2020cnn}. Predictions are made for 3 years (2016-2018) where as with the remote sensing data, reporting the RMSE averaged over 3 runs of 3 seeds, the baseline performance refers to traditional training of a single global dataset, whilst the model sharing ensemble and federated learning procedures are trained on the local subsets of data belonging to each state. 

\begin{figure}
    \captionsetup[subfigure]{font=small,labelfont=small}
    \centering
    \begin{subfigure}[t]{0.49\textwidth}
      \centering
        \includegraphics[width=0.95\textwidth]{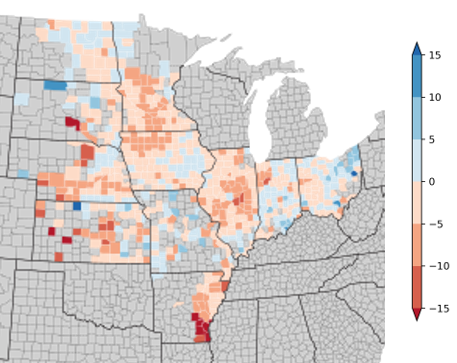}
        \caption{Baseline.}
    \end{subfigure}
    \hfill
    \begin{subfigure}[t]{0.49\textwidth}
      \centering
        \includegraphics[width=0.95\textwidth]{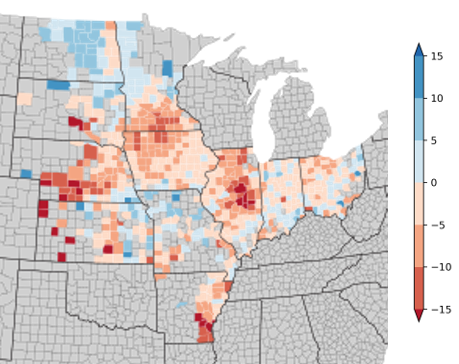}
        \caption{Ensemble of Models.}
    \end{subfigure}
    
    \begin{subfigure}[t]{0.49\textwidth}
      \centering
        \includegraphics[width=0.95\textwidth]{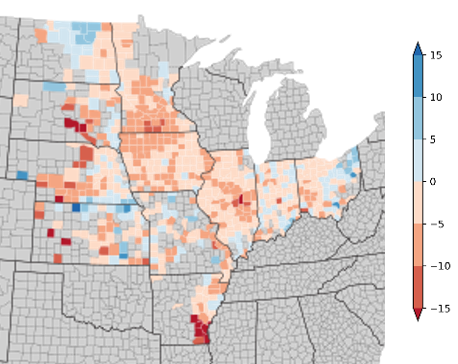}
        \caption{Federated Learning (FedBN).}
    \end{subfigure}%
    \hfill
    \begin{subfigure}[t]{0.49\textwidth}
      \centering
        \includegraphics[width=0.95\textwidth]{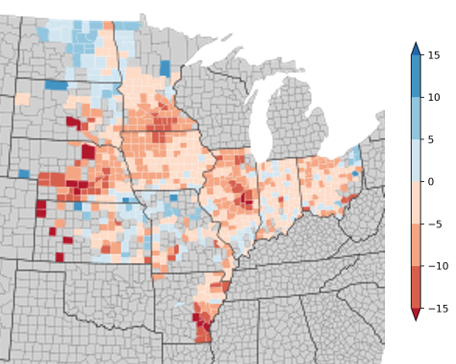}
        \caption{Federated Learning + LDP.}
    \end{subfigure}
    \caption{State-level visualization of the CNN network under each training procedure for the remote sensing data, RMSE (Bushels per Acre) per county prediction for year 2015. Color bar represents difference in RMSE (Bushels per Acre) from ground truth value.}
    \label{fig:sat_image}
\end{figure}

\subsection{Model Sharing Performance} \label{ms_per}

The model sharing ensemble procedure defined in Section \ref{modelsharing} trains a model locally on a single silo of data pertaining to a single state. At test time, inference is performed on the models from each state computing a prediction of yield for that state, these predictions are averaged over all the individual models trained on each local data silo. This methodology follows our initial proposition most simply as to share information captured by the machine learning models rather than the data itself. Such an approach is empirically shown in Table~\ref{tab:sat_cnn}, \ref{tab:sat_lstm} and \ref{tab:tab} to perform well across all modalities in the task of crop yield prediction with an approximate 1.32\% and 9.09\% increase in RMSE for the image and tabular modalities respectively from local baselines.  

To improve the ensemble method extending from a simple average we implement a simple distance rank to introduce a weighted average of models as to most appropriately leverage models trained on data that originate in geographical proximity. As defined in \cite{you2017deep} the influence on geographic location between the data plays significant impact on performance due to climate, soil and weather changes, as such our distance weighting scheme aims to contribute to the same problem. Although a small improvement, Table~\ref{tab:Distance_weight} shows how an approximately 0.2 and 1.5 RMSE reduction for the image and tabular modalities respectively can be found on average across all years with the addition of a simple weighting scheme. All results given for the model sharing ensembles utilize this weighting scheme unless stated otherwise.

When making comparisons of the model sharing approach we first look at the traditional training and local model training baseline. Shown in Table~\ref{tab:sat_cnn}, \ref{tab:sat_lstm} and \ref{tab:tab} we observe how the model sharing approach makes improvements over the local model training procedure (models are trained on a local silo and evaluated only on local data only, i.e. state 1 model is trained and evaluated on state 1 test data only) with a 0.3 reduction in RMSE in the tabular modality showing how the introduction of ensembles can improve predictions on local data. We conjecture this improvement is attributed to the variability in the sources and as such the average prediction represents a more generalized view across state conditions. However, when comparing to traditional learning baseline we observe the expected but significant performance derogation. This can primarily be attributed to quantity and variability of data present in each data silo, where overfitting observed for each local model during training with identical model architecture (to maintain fair experimentation). This is particularly relevant to the remote sensing data where the datasets per silo are comparatively small, in the tabular modality we observe a less significant drop in performance due to the larger dataset size. To further evaluate this hypothesis on the remote sensing dataset, we reduce the number of silos to 4 and therefore increasing the size of each local silo dataset, subsequently we see some improvement to an average 6.023 RMSE for the CNN model, yet this is still limited by quantity of data. 

Although in our test cases the ensemble of shared models performs well but not ideally, the setting of larger datasets per silos can be a contributing factor to the success of such a training paradigm in training individual models. We see the simplicity of ensembles to be a desirable trait in facilitating data sharing, whilst settings with larger local datasets performance can be expected to lie closer to traditionally pooled baselines whilst significantly outperforming local training.

\begin{table}[h]
    \scriptsize
    \centering
    \begin{tabular}{c|c|c}
        Weighting & Imaging Dataset & Tabular Dataset \\
        \midrule
        None & 6.716 & 3.714 \\
        Distance Rank & 6.544 & 3.885 \\
    \end{tabular}
    \caption{Comparison of ensemble state distance rank weighting scheme under both dataset modalities, average RMSE (Bushels per Acre) over all prediction years (CNN model for remote sensing dataset).}
    \label{tab:Distance_weight}
\end{table}

\subsection{Cross-Silo Federated Learning Performance} \label{FL_per}

To address the limitations observed in training many local models solely on local data (e.g. reduced variability, difficulty in training small datasets due to overfitting, etc.), we proposed the use of federated learning (Section~\ref{federated}) which trains a single global model via a series of local model updates and aggregations. Federated learning leverages all data from all silos via the aggregation of model updates, and resultantly produces a model that is not only effectively trained on the union of the individual datasets simultaneously, but also captures the variability in the model from this union. To test the performance we evaluate the final aggregated global model on each of the individual test sets from each silo. 

Table~\ref{tab:sat_cnn}, \ref{tab:sat_lstm} and \ref{tab:tab}, including Figure~\ref{fig:sat_image} show the performance of the federated learning training method alongside the baseline approaches of traditional and local model training. We observe across all prediction years, model architectures and data modality how federated learning methods outperform the local training baseline and ensemble of shared model approaches, demonstrating an approximate 22.75\% and 39.81\% improvement in RMSE to local baselines for the image and tabular modalities respectively. More importantly, we observe how federated learning training performs nearly identically to traditional training baseline, with a small but expected disparity in performance, 6.68\%. The ability to effectively train machine learning models in a distributed and independent nature without the disclosure of raw data is a vastly important observation in the agri-food sector and can lead to facilitating collaboration of multiple parties with reduced concern of performance reduction. 

Furthermore, the implementation of federated learning systems can be greatly scaled to many situations involving a large number of clients or, as demonstrated here, with few clients (11, and 9 silos). Our work employs the use of the federated batch normalization aggregation procedure to help reduce the effect of statistical heterogeneity between local datasets Table~\ref{tab:FedMethods}. Although not applicable in every task, we demonstrate how the careful consideration of algorithmic choice can not only achieve greater performance ( Table~\ref{tab:FedMethods} ) but also reduce computational burdens by reducing time to reach convergence as shown in Figure \ref{fig:rounds}. Although, the nuances displayed by particular model architectures, datasets and tasks introduce very application specific problems, and must be address case-by-case, the proposition of federated learning in agri-food provides a empirically and theoretically sound basis for collaboration.

\begin{table}[h]
    \centering
    \scriptsize
    \begin{tabular}{c|c|c|c|c|c|c}
        \multicolumn{7}{c}{CNN-RNN} \\
        Year & \makecell{Traditional \\Baseline} & \makecell{Local \\Baseline} & \makecell{Model \\Sharing} & \makecell{Model \\Sharing + LDP} & \makecell{Federated \\Learning} & \makecell{Federated \\Learning + LDP} \\
        \midrule
        2009 & 4.735 & 5.684 & 6.774 & 7.033 & 5.013 & 6.862 \\
        2010 & 5.167 & 7.076 & 6.667 & 7.969 & 5.691 & 6.970 \\
        2011 & 6.009 & 6.606 & 6.915 & 8.480 & 5.859 & 7.103 \\
        2012 & 5.968 & 7.605 & 6.747 & 8.401 & 6.235 & 7.345 \\
        2013 & 5.246 & 6.936 & 6.251 & 6.954 & 5.352 & 6.256 \\
        2014 & 4.915 & 6.173 & 5.960 & 6.869 & 5.017 & 6.561 \\
        2015  & 5.073 & 6.337 & 6.495 & 8.001 & 5.981 & 6.683 \\
        \midrule
        Avg & 5.302 & 6.631 & 6.544 & 7.672 & 5.592 & 6.825 \\
        
    \end{tabular}
    \caption{Remote sensing image dataset: RMSE (Bushels per Acre) of county-level performance under each training procedure for the CNN model. The values reported are the average of 3 runs of 3 random initialization seeds. For the LDP variants $\epsilon=8$ and $\delta=1\times10^{-5}$.}
    \label{tab:sat_cnn}
\end{table}

\begin{table}[h]
    \centering
    \scriptsize
    \begin{tabular}{c|c|c|c|c|c|c}
        \multicolumn{7}{c}{CNN} \\
        Year & \makecell{Traditional \\Baseline} & \makecell{Local \\Baseline} & \makecell{Model \\Sharing} & \makecell{Model \\Sharing + LDP} & \makecell{Federated \\Learning} & \makecell{Federated \\Learning + LDP} \\
        \midrule
        2009 & 5.059 & 6.972 & 6.972 & 8.374 & 6.556 & 7.077\\
        2010 & 5.710 & 7.643 & 6.768 & 7.381 & 6.045 & 7.721\\
        2011 & 6.560 & 7.525 & 7.402 & 7.853 & 6.849 & 7.123\\
        2012 & 7.262 & 8.164 & 7.512 & 8.185 & 6.641 & 6.564\\
        2013 & 5.344 & 8.030 & 6.831 & 7.704 & 5.617 & 6.650\\
        2014 & 5.465 & 7.457 & 6.571 & 7.137 & 5.175 & 5.708\\
        2015 & 6.235 & 7.823 & 6.894 & 7.404 & 5.774 & 6.631\\
        \midrule
        Avg & 5.948 & 7.659 & 6.993 & 7.719 & 6.094 & 6.782\\
        
    \end{tabular}
    \caption{Remote sensing image dataset: RMSE (Bushels per Acre) of county-level performance under each training procedure for the LSTM model. The values reported are the average of 3 runs of 3 random initialization seeds. For the LDP variants $\epsilon=8$ and $\delta=1\times10^{-5}$.}
    \label{tab:sat_lstm}
\end{table}

\begin{table}[h]
    \centering
    \scriptsize
    \begin{tabular}{c|c|c|c|c|c|c}
        \multicolumn{7}{c}{LSTM} \\
        Year & \makecell{Traditional \\Baseline} & \makecell{Local \\Baseline} & \makecell{Model \\Sharing} & \makecell{Model \\Sharing + LDP} & \makecell{Federated \\Learning} & \makecell{Federated \\Learning + LDP} \\
        \midrule
        2016 & 2.601 & 4.819 & 4.325 & 4.854 & 2.969 & 4.044 \\
        2017 & 2.879 & 4.529 & 4.283 & 4.537 & 2.931 & 3.561 \\
        2018 & 2.326 & 3.148 & 3.048 & 3.317 & 2.475 & 2.572 \\
        \midrule
        Avg & 2.602 & 4.165 & 3.885 & 4.236 & 2.782 & 3.392 \\
        
    \end{tabular}
    \caption{Tabular weather, soil and crop management dataset: RMSE (Bushels per Acre) of county-level performance under each training procedure for the CNN-RNN model. The values reported are the average of 3 runs of 3 random initialization seeds. For the LDP variants $\epsilon=1.5$ and $\delta=1\times10^{-7}$.}
    \label{tab:tab}
\end{table}

\subsection{Differential Privacy Performance} \label{FL_per}

The core functionality of our proposed methods to facilitate information sharing have demonstrated empirically their ability to perform close to the traditional training baseline and outperform local non-collaborative training. As described in Section \ref{localdiff}, we employ differential privacy at local/client level to mitigate such attacks. Furthermore, the implementation of differential privacy is a privacy-performance trade-off defined by the practitioner, where in our LDP setting we employ $(\epsilon, \delta)$-LDP in which $\epsilon$ and $\delta$ vary for a given task, and dataset. It is important to emphasize that the desired value of $\epsilon$ and $\delta$ are vastly dependent on the data and resulting trade-off between privacy and performance for a given task. The $\epsilon$ value --- otherwise known as the privacy budget --- is defined as the maximum distance between a query on the dataset $d$ and the same query on dataset $d'$. Thus, when the distance is small an adversary may be unable to determine which dataset a value originated from given the small distance between the two sets, this is observed mathematically in Equation~\ref{eq:dp}. 

\begin{figure}[h]
    \centering
    \includegraphics[width=0.64\textwidth]{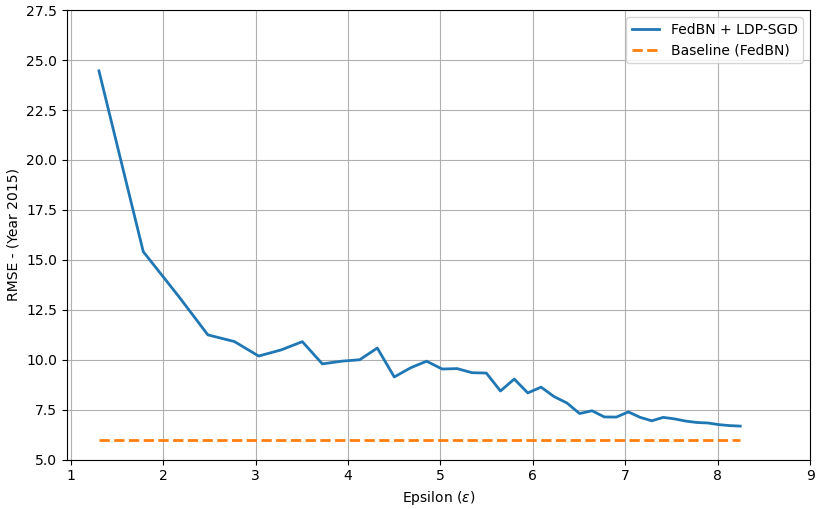}
    \caption{Effect of RMSE (Bushels per Acre) for predict year 2015 for the CNN model trained on the remote sensing dataset as LPD privacy budget $\epsilon$ increases.}
    \label{fig:epsilon}
\end{figure}

Regarding the remote sensing dataset, given the aforementioned limitation of data quantity the privacy guarantee is also reduced~\cite{geyer2017differentially}, we train our federated and models sharing ensemble networks under the LDP-SGD procedure with a noise value of 1.4 and gradients clipped that have a norm greater than 12 , this value had been defined from the median gradients of the network. The training was terminated once $\epsilon=8.0$ whilst keeping $\delta=1\times10^{-5}$, the results of this setup are given in Table~\ref{tab:sat_cnn} and Table~\ref{tab:sat_lstm}. We select $\epsilon=8$ empirically as the point that performance privacy trade-off is roughly converging (Figure~\ref{fig:epsilon}), and as such gave strong performance under federated learning with a 19.85\% and 10.68\% reduction in RMSE for CNN and LSTM models respectively. 

Under the tabular modality dataset we employ an identical noise value and reduce the gradient clipping to 10 given different median values for the CNN-RNN network. Furthermore, we utilize a smaller $\delta=1\times10^{-7}$ given the greater samples in the dataset, and as a result our value for $\epsilon$ is correspondingly lower, $\epsilon=1.5$, which is when we terminate training. As with the remote sensing data we observe a similar 19.76\% reduction in RMSE under the federated learning procedure  with the addition of differential privacy. This is expected due to $\epsilon$ being empirically selected as the initial point before convergence. Across both datasets we observe that ensemble of model sharing with differential private mechanisms performs comparatively poorly, this can be mainly attributed to the observations of model sharing that such a method is limited by dataset size and feature heterogeneity between local silos given that the reduction in RMSE is aligned with the reduction in the federated learning case.%

\section{Discussion}\label{discussion}
Thus far in this paper, we have conducted an empirical demonstration of how data sharing can be enabled through distributed and data independent machine learning training, maintaining an adaptable level of privacy with the hope to showcase how collaborative learning can be achieved without disclosure of commercially sensitive information. Our example implementations address an established and highly sought after problem of production optimization in a distributed setting for two modalities of data, images and tabular samples. This is just one potential use case that aims to demonstrate how a federation or consortium of individual actors can leverage shared data to improve production optimization, there exist many other potential benefits as described in Section \ref{usecases} that help achieve improved food safety and sustainability to meet a variety of regulatory requirements \cite{EUfarmtofork,UKpathtosustainable}.

Firstly, our empirical analysis demonstrates the applicability of such methods for appropriate in agri-food production optimization machine learning tasks. Specifically, we show that the federated setting under both dataset modalities performs exceptionally well achieving results that are considered extremely close to traditional machine learning baselines with a 5.32\% and 6.68\% reduction in RMSE for the image and tabular dataset respectively. Significantly, we also show that our proposed methods outperform the locally trained models further solidifying our proposition that data or information sharing can benefit all parties involved to produce better performing models. We based our comparison and definition of adequate performance on the original baseline papers for each corresponding dataset \cite{you2017deep,khaki2020cnn}, and our replicating baseline models on identical datasets with some small alterations to the networks defined in Section~\ref{experimental}. Across both datasets we achieve greater performance than reported in both original works, and our federated setting performs better than or around the reported values. We consider these results to be a great success demonstrating the potential of these data independent methods for training on complex dataset. On the other hand, we also observe that our ensemble of shared models in the remote imaging setting does not perform in a threshold we would consider ideal, this can be primarily attributed to the dataset size and difficulty of training small sets with an over parameterized model (model architectures remained the same for direct comparison of training procedure) as explained previously. Additionally, we conjecture that the ensemble of shared models' performance deficit can also be attributed to the statistical heterogeneity between local data silos and as such when testing on a different silos test data the trained models may have difficulty addressing the present feature shift.

Vitally, our methods aim to elicit confidence and trust in participants via privacy preservation techniques that demonstrate theoretic guarantees to maintain privacy. Most obviously is the maintenance of data independence and removal of raw data sharing, whilst still being able to collaboratively train machine learning models. Moreover, we address the concerns around the malicious attempts to obtain training data from the shared information, a significant implication regarding commercial sensitivity among participants. The nature of distributed training allows for additional technologies to be used in tandem to help introduce more stringent privacy measures. In practice there exist other privacy preserving machine learning approaches, such as fully-homomorphic encryption~\cite{onoufriou2021fully}, that could facilitate more confidence in data sharing. 

As previously alluded when defining our methods, there still exist social and political challenges in adoption of distributed training for data sharing. Most notably, the coordination of training on individual participant data and distribution of trained models to participants may introduce concerns of competitive advantage from the coordinating party. 
Our work provides processes that solve the core social implications of trust in sharing raw data further removing this necessity, yet the understanding of the processes by the participants is they driving factor to changing perspectives.

\section{Conclusion}
Many of the agri-food sector's implications involving transparency and holistic data analysis stem from technological lagging, and hostility to data sharing \cite{sarpong2014traceability,ODIfoodwaste} that can inevitably lead to difficulties meeting many of the mandatory requirements in an efficient manner. We proposed the use of distributed, machine learning training procedures to overcome the strong social barriers relating to commercial sensitivity and unwillingness to share raw data. To the best of our knowledge this is the first time this type of machine learning setting, specifically federated learning, is explored in the agri-food sector which could potentially benefit the industry and consumers as a whole. 

We give theoretical descriptions and empirical evidence that the proposed methods will not only provide privacy guarantees supporting the facilitation of data sharing and collaborative machine learning training, but also perform close to their traditionally trained machine learning counterparts. The performance demonstrated gives further credibility to such methods for training machine learning models in a collaborative setting providing knowledge that performance can be guaranteed and improved with more data from differing sources. 

The field of distributed and privacy preserving machine learning is growing rapidly, where these methods are becoming evermore relevant in industry, we believe this work is just the start point for the adoption of large scale distributed computation in agri-food. Direct future work aims to showcase the implementation of these propositions in the real-world setting whilst further addressing the issues regarding statistical heterogeneity as well as addressing the social, political challenges present in data sharing. Moving forward we aim to further develop a more concrete pipeline from data to model output addressing communication, central servers and regulatory implications to achieve a standard in collaborative and data sharing procedures across the food supply chain. 
\section*{Acknowledgements}
This work was supported by an award made by the UKRI/EPSRC funded Internet of Food ThingsNetwork+ grant EP/R045127/1.

\newpage

\end{document}